\documentclass{article}

\usepackage{microtype}
\usepackage{graphicx}
\usepackage{subfigure}
\usepackage{booktabs} 
\usepackage{subcaption}
\usepackage{makecell}
\usepackage{algorithm}
\usepackage{algorithmic}
\usepackage{diagbox}
\usepackage{multirow}

\usepackage{hyperref}

\usepackage[accepted]{icml2024}

\usepackage{amsmath}
\usepackage{amssymb}
\usepackage{mathtools}
\usepackage{amsthm}
\usepackage{amsfonts}

\let\svthefootnote\thefootnote
\newcommand\freefootnote[1]{%
  \let\thefootnote\relax%
  \footnotetext{#1}%
  \let\thefootnote\svthefootnote%
}

\usepackage[capitalize,noabbrev]{cleveref}

\theoremstyle{plain}

\theoremstyle{definition}

\theoremstyle{remark}

\ifdefined\isarxiv
\usepackage{amsmath}
\usepackage{amssymb}
\usepackage{mathtools}
\usepackage{amsthm}
\usepackage{subcaption}
\usepackage{makecell}
\usepackage{algorithm}
\usepackage{algorithmic}
\usepackage{diagbox}
\usepackage{hyperref}
\usepackage{microtype}
\usepackage{graphicx}
\usepackage{subfigure}
\usepackage{booktabs}
\fi
\ifdefined\isarxiv
\else
\usepackage[textsize=tiny]{todonotes}

\icmltitlerunning{Improving Prototypical Visual Explanations with Reward Reweighing, Reselection, and Retraining}
\fi

\begin{document}

\twocolumn[
\icmltitle{Improving Prototypical Visual Explanations with \\ Reward Reweighing, Reselection, and Retraining}



\icmlsetsymbol{equal}{*}

\begin{icmlauthorlist}

\icmlauthor{Aaron J. Li}{harvard}
\icmlauthor{Robin Netzorg}{berkeley}
\icmlauthor{Zhihan Cheng}{berkeley}
\icmlauthor{Zhuoqin Zhang}{berkeley}
\icmlauthor{Bin Yu}{berkeley}
\end{icmlauthorlist}

\icmlaffiliation{harvard}{Harvard University}
\icmlaffiliation{berkeley}{University of California, Berkeley}

\icmlcorrespondingauthor{Aaron J. Li}{jiaxun\_li@g.harvard.edu}
\icmlcorrespondingauthor{Bin Yu}{binyu@berkeley.edu}

\icmlkeywords{Machine Learning, Interpretability, Human-in-the-loop}

\vskip 0.3in
]

\printAffiliationsAndNotice{}  

\begin{abstract}
In recent years, work has gone into developing deep interpretable methods for image classification that clearly attributes a model's output to specific features of the data. One such of these methods is the \textit{prototypical part network} (ProtoPNet), which attempts to classify images based on meaningful parts of the input. While this architecture is able to produce visually interpretable classifications, it often learns to classify based on parts of the image that are not semantically meaningful. To address this problem, we propose the \textit{reward reweighing, reselecting, and retraining} (R3) post-processing framework, which performs three additional corrective updates to a pretrained ProtoPNet in an offline and efficient manner. The first two steps involve learning a reward model based on collected human feedback and then aligning the prototypes with human preferences. The final step is retraining, which realigns the base features and the classifier layer of the original model with the updated prototypes. We find that our R3 framework consistently improves both the interpretability and the predictive accuracy of ProtoPNet and its variants. 
\end{abstract}

\section{Introduction}
With the widespread use of deep learning, making large models interpretable has become an increasingly important goal for the machine learning community. As these models continue to see use in high-stakes situations, practitioners hoping to justify a decision need to understand how a deep model makes a prediction, and trust that those explanations are valuable and correct \citep{rudin2021interpretable}. One such proposed method for image classification is the \textit{prototypical part network} (ProtoPNet), which classifies a given image based on its similarities to prototypical parts of different classes \citep{chen2019looks}. This model aims to combine the power of deep learning with an intuitive reasoning module similar to humans. 

While ProtoPNet aims to learn meaningful prototypical concepts, in practice, learned prototypes suffer from learning spurious features, such as the background of an image, and inconsistent concepts, such as learning both the head and the wing of a bird \citep{bontempelli2023conceptlevel}. Problems like these are highly detrimental to the transparency and efficacy of these models, and thus make the models less likely to be utilized by a human user. Various methods have been proposed to improve such questionable visual explanations \citep{Nauta_2021, barnett2021iaiabl, bontempelli2023conceptlevel, evalprotopnet, ma2023looks}, but none of them have attempted to explicitly quantify the human user's preference for the prototypes. 

Thus, in addition to improve the model performance itself, the main goal of this work is to prompt the model to produce prototypes that are more aligned with human preferences, which is a crucial step towards model interpretability \citep{human-interpretability}. These two objectives also correspond to the \textit{predictive accuracy} and \textit{relevancy} within the well-known predictive,descriptive, relevant (PDR) interpretable machine learning framework \citep{murdoch2019definitions}.

\freefootnote{\\The source code of this work is available at \url{https://github.com/aaron-jx-li/R3-ProtoPNet}.}

Towards this end, we propose the \textit{reward reweighing, reselecting, and retraining} (R3) concept-level debugging framework which improves the original ProtoPNet by using a learned reward model to improve the quality of the prototypes. Our method doesn't need to train the model from scratch, and we call the debugged model R3-ProtoPNet. The human feedback R3 requires is a small number of rating data of prototype quality with multiple scales, given by users when they are shown visualizations of image-prototype pairs. With limited human feedback data on the Caltech-UCSD Birds-200-2011 (CUB-200-211) dataset \citep{cubbirbs}, we are able to train a high-quality reward model that achieves $90.1\%$ test accuracy when ranking human preferences, serving as a strong measure for prototype quality. Two distinct advantages of having an external reward model that faithfully captures human preferences are: 

\begin{itemize}
    \item The debugging process becomes efficient, because the reward model is pretrained and it doesn't require online feedback on the explanations generated by the current model.
    \item The metric for prototype quality becomes more maneuverable, as different reward models could capture slightly different user preferences. 
\end{itemize}

We train this reward model from a small pairwise human preference dataset (further explained in sections \ref{human_collection} and \ref{reward-learning}). Then, the R3 framework evaluates and updates the prototypes to maximize their induced rewards, and these debugging steps are followed by a retraining step to restore the predictive performance. Empirically, the R3 debugging procedure is able to reduce model's dependence on spurious features and make the visual explanations more favorable to users. When used either individually or as base learners in an ensemble, R3-ProtoPNet outperforms the original ProtoPNet on a held-out test dataset in terms of predictive accuracy. In general, our proposed framework improves upon a class of widely used inherently interpretable deep learning models (i.e. prototype-based models) by efficiently utilizing their own interpretations.

The contributions of this work can be summarized as follows: 

\begin{itemize}
\item We propose using the learned reward model as a quantified metric of prototypical visual explanation quality and model interpretability
\item We introduce the R3 framework and R3-ProtoPNet, which use efficient reward-guided debugging to improve both prototype meaningfulness and predictive performance. 
\end{itemize}

\section{Related Work}

\subsection{Example-based Models and Prototypical Part Networks}

 There are many explainability and interpretability methods available to the user within the field of interpretable machine learning \citep{rudin2021interpretable}, and the two main goals for the community are (1) to come up with inherently interpretable machine learning paradigms \citep{agarwal2022hierarchical} and (2) to propose reliable explanation methods for model outputs \citep{ribeiro2016should, lundberg2017unified, murdoch2018beyond}. To ground the discussion, we focus primarily on example-based models, one such example being ProtoPNet. While other example-based methods exist, such as the non-parametric xDNN \citep{angelov2019explainable} or SITE \citep{wang2021selfinterpretable}, which performs predictions directly from interpretable prototypes, we focus on the ProtoPNet due to its intuitive reasoning structure and explicit visual explanations.

Since ProtoPNet's first introduced by \citet{chen2019looks}, many iterations of follow-up works have been proposed. Work has explored extending the ProtoPNet to different architectures such as transformers \citep{xue2022protopformer}, or sharing class information between prototypes \citep{protopshare}. \citet{Donnelly_2022_CVPR} increase the spatial flexibility of ProtoPNet, allowing prototypes to change spatial positions depending on the pose information available in the image. ProtoPNets and variations have seen success in high-stakes applications, such as kidney stone identification \citep{floresaraiza2022interpretable} and mammography \citep{barnett2021iaiabl}.

Many works have also worked on addressing the original ProtoPNet's overemphasis on spurious features. \citet{Nauta_2021} introduce an explainability interface to ProtoPNet, allowing users to see the dependence of the prototype on certain image attributes. \citet{barnett2021iaiabl} introduce a variation of the ProtoPNet, IAIA-BL, which biases prototypes towards expert labelled annotations of classification-relevant parts of the image. Other works such as \citet{evalprotopnet} and \citet{ma2023looks} incorporate new modules and constraints into ProtoPNet to improve its empirical performance without using human feedback. 

Similar to how we provide human feedback at the interpretation level, \citet{bontempelli2023conceptlevel} introduce ProtoPDebug, which first asks for binary user feedback on prototypes as "forbidden" or "valid", and then uses a fine-tuning step that includes the collected feedback as a supervised constraint into the ProtoPNet loss function. 

Compared with previous approaches, our R3 framework allows users to efficiently collect high-quality human feedback data and train a robust reward model that could be used to both evaluate and debug the original prototypes. 

\subsection{Learning from Human Feedback}

As the term interpretability lacks a mathematical quantification, practitioners have argued that evaluating it well requires human feedback \citep{rigorous17}. Our method starts from learning a reward model from human feedback, and then use it as an interpretability measure.

Since the success of InstructGPT \citep{ouyang2022training}, Reinforcement Learning from Human Feedback (RLHF) has attracted a lot of attention. But before that, incorporating human feedback into reinforcement learning methods via a learned reward model also has a deep history in reward learning \citep{christiano2017deep, jeonrewarddesign}. Some prior works incorporate the reward function as a way to weigh the likelihood term \citep{stiennon2022learning, ziegler2019fine}. While works taking inspiration from InstructGPT have used proximal policy optimization (PPO) to fine-tune networks with human feedback \citep{bai2022training}, it is unclear to the extent that formal reinforcement learning is necessary to improve models via learned reward functions \citep{lee2023aligning}, or if the human feedback needs to follow a particular form \citep{askell2021general}.  

Different from RLHF, our work doesn't rely on any formal RL algorithms, and instead simply uses reward values as a supervisory signal to guide the search of semantically meaningful prototypes.

\section{Reward Reweighed, Reselected, and Retrained Prototypical Part Network (R3-ProtoPNet)} \label{rwr} 

In this section, we first describe the basics of ProtoPNet \citep{chen2019looks}, and then present our R3 debugging framework in detail, which includes the collection of high-quality human feedback data, our reward model, and the incorporation of the reward model into debugging via a three-step update procedure.  

\begin{algorithm}
\caption{Reward Reweighed, Reselected, and Retrained Prototypical Part Network (R3-ProtoPNet)}\label{alg:R3-ProtoPNet}

\begin{algorithmic}[1]
\STATE \textbf{Initialize:} Collect high-quality human feedback data and train a reward model.

\STATE \textbf{Reward Reweighing:} Perform the reward-reweighed update for the ProtoPNet, defined in Equation \ref{tab:eq1}.
Optimize the loss function, which leads to locally maximal solutions, improving the prototypes. 

\STATE \textbf{Prototype Reselection:} Run the reselection procedure based on a reward threshold. \\
If $\frac{1}{n_k}\sum_{i\in I(p_j)}r(x_i, p_j) < \alpha$, reselect the prototype by sampling from patch candidates and temporarily setting the prototype to a new candidate that passes the acceptance threshold and is unique from other current prototypes.

\STATE \textbf{Retraining:} Retrain the model with the same loss function used in the original ProtoPNet update, to realign the prototypes and the rest of the model.
\end{algorithmic}
\end{algorithm}

\subsection{Preliminaries on ProtoPNet}

Here we adopt the notation used in \citet{chen2019looks}. The ProtoPNet architecture builds on a base convolutional neural network $f$, which is then followed by a prototype layer denoted $g_p$, and a fully connected layer $h$. Typically, the convolutional features are taken pretrained models like VGG-19, ResNet-34, or DenseNet-121.

The ProtoPNet injects interpretability into these convolutional architectures with the prototype layer $g_p$, consisting of $m$ prototypes $\boldsymbol{P} = \{p_j\}^{m}_{j=1}$ typically of size $1\times1\times D$, where $D$ is the shape of the convolutional output $f(x)$. By keeping the depth the same as the output of the convolutional layer, but restricting the height and width to be smaller than that of the convolutional output, the learned prototypes select a patch of the convolutional output. Reversing the convolution leads to recovering a prototypical patch of the original input image $x$. Using upsampling, the method constructs an activation pattern per prototype $p_j$. 

To use the prototypes to make a classification given a convolutional output $z=f(x)$, ProtoPNet's prototype layer computes a max pooling over similarity scores: $g_{p_j}(z) = \max_{\Tilde{z}\in\text{patches}(z)}\log(({\|\Tilde{z} - p_j\|}_2^2 + 1)({\|\Tilde{z} - p_j\|}_2^2 + \epsilon))$, for some small $\epsilon < 1$. This function is monotonically decreasing with respect to the distance, with small values of ${\|\Tilde{z} - p_j\|}_2^2$ resulting in a large similarity score $g_{p_j}(z)$. Assigning $m_k$ prototypes for all $K$ classes, such that $\sum_{k=1}^K m_k = m$, the prototype layer outputs a vector of similarity scores that matches parts of the latent representation $z$ to prototypical patches across all classes. The final layer in the model is a linear layer connecting similarities to class predictions.

In order to ensure that the prototypes match specific parts of training images, during training the prototype vectors are projected onto the closest patch in the training set. For the final trained ProtoPNet, every $p_j$ corresponds to some patch of a particular image. 

\subsection{Human Feedback Collection}\label{human_collection}

As mentioned earlier, while ProtoPNet is capable of providing interpretable classifications, the naive training described in \citep{chen2019looks} results in prototypes that focus on spurious and inconsistent features \citep{barnett2021iaiabl, bontempelli2023conceptlevel}. 

A crucial aspect behind the success of learning faithful reward models is the collection of high quality human feedback data. Unclear or homogeneous feedback may result in a poor performing reward model \citep{christiano2017deep}. The design of human feedback collection is vitally important to the training of a useful reward model.

The inherent interpretability of ProtoPNet is particularly useful for reward learning. Given a trained ProtoPNet, it is possible for a user, who doesn't have to be an expert, to directly critique the learned prototypes. In the case of classifying birds in the CUB-200-2011 dataset, it is clear that if a prototype gives too much weight to the background of the image (spurious), or if the prototype corresponds to different parts of the bird when looking at different images (inconsistent). Given these prototypes that fail to contribute to prediction, a lay person trying to classify birds would be able to rate these prototypes as "bad", with a proper rating rubric. 

\begin{figure}[t]
  \includegraphics[width=0.5\textwidth, trim = 4.5cm 18cm 3.5cm 1cm]{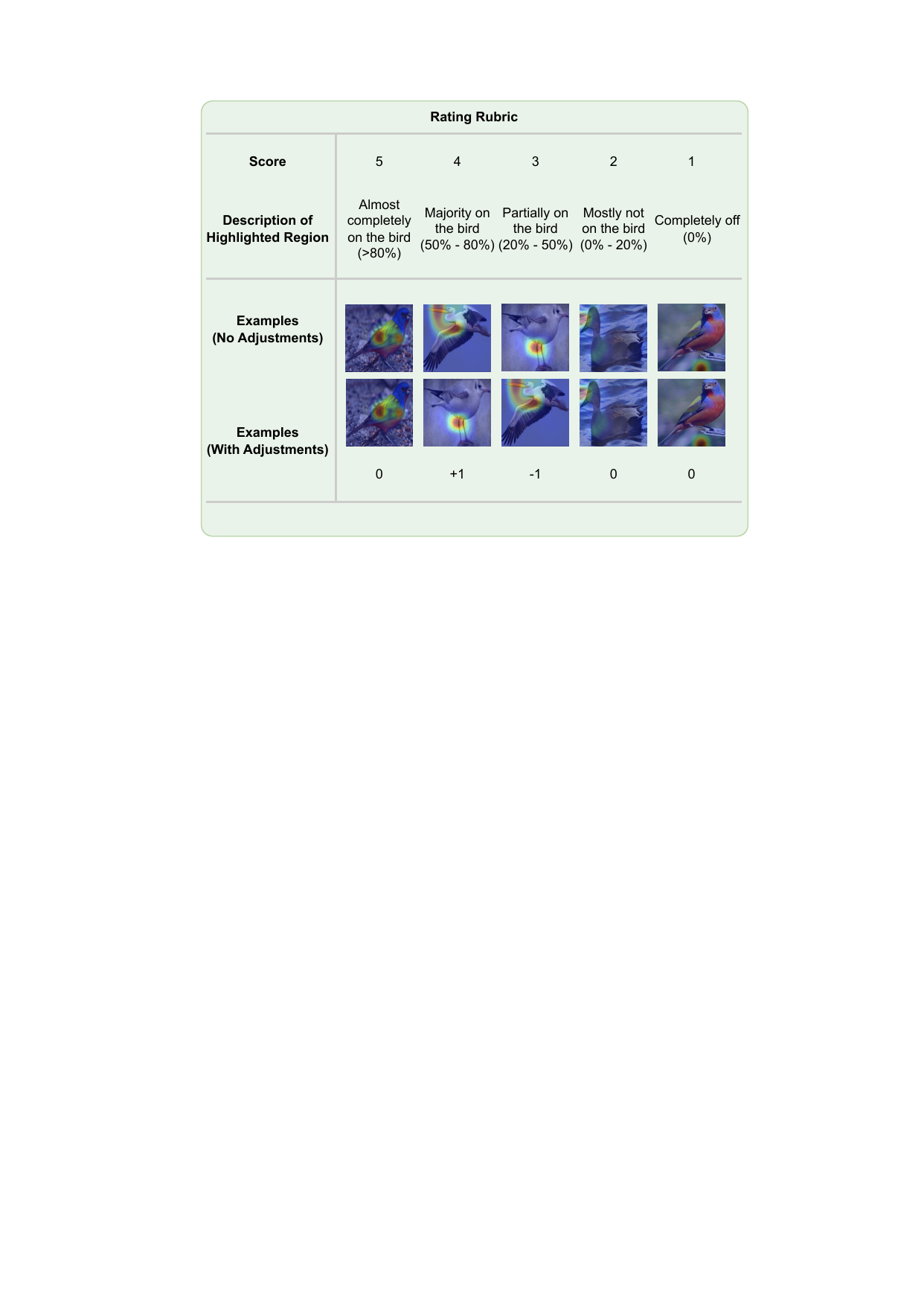}
\caption{Rubric used for human feedback on the activation patterns of predictions for birds from the CUB-200-2011 dataset. First, the rater estimates a base score based on overlap proportion, and then an optional adjustment $\delta \in \{-1, 1\}$ could be given based on how meaningful or characteristic the focused body part is.} \label{fig:scale}
\end{figure}

There are many different ways to elicit the preference of a user \citep{askell2021general}. Although it is possible to incorporate many different forms of feedback into the R3-ProtoPNet, such as asking a user to compare prototypes to elicit preferences or ask for a binary value of whether a prototype is "good" or "bad", we found most success with asking the users to rate a prototype on a scale from 1 to 5. While scalar ratings can be unstable across different raters, with a clear, rule-based rating method, rating variance is reduced and it is possible to generate high-quality labels. An example rating scale on the CUB-200-2011 dataset is provided in Figure \ref{fig:scale}.

\subsection{Reward Learning}
\label{reward-learning}

We note that, when a user provides feedback on a prototype, it is not the training image or the model prediction that the user is providing feedback on, but the prototype's resulting interpretation: the activation patterns. Our task is therefore different from RLHF applied to language modeling or RL tasks \citep{ouyang2022training, christiano2017deep}, where human feedback is provided on the model output or resulting state. We therefore collect a rating dataset $\mathcal{D} = \{(x_i, y_i, h_{i,j}, r_{i,j})\}_{i=1,j=1}^{n,m}$, where $x_i,y_i$ are the training image and label, and $h_{i,j},r_{i,j}$ are prototype $p_j$'s activation patterns and user-provided ratings for image $x_i$. We note that collecting preferences for this entire dataset is prohibitive and unnecessary, so we only collect a subset. 

 Given the dataset $\mathcal{D}$, we generate the induced comparison dataset, whereby each entry in $\mathcal{D}$ is paired with one another. Given $i\neq i'$ and/or $j\neq j'$, we populate a new paired dataset, $\mathcal{D}_{paired}$, which consists of the entries of $\mathcal{D}$ indexed by $i,j,i',j'$, and a comparison $c$, which takes values $-1, 1$. If the left-hand sample is greater, and therefore considered higher-quality, $r_{i,j} > r_{i',j'}$, then $c = -1$. If the right-hand sample is greater $r_{i,j} < r_{i',j'}$, then $c = 1$. This synthetic construction allows us to model the reward function, $r(x_i, h_{i,j})$ via the Bradley-Terry Model \citep{bradley1952rank}, which has demonstrated success in learning pairwise user preferences \citep{christiano2017deep}. We train this model with the same loss function as in \citet{christiano2017deep}, a cross-entropy loss over the probabilities of ranking one pair over the other (See Appendix \ref{tab:pairwise-loss} for details). This synthetic construction combinatorially increases the amount of preference data, allowing us to train a high-quality reward model on relatively small amounts of human feedback data.

\subsection{Reward Reweighed, Reselected, and Retrained Prototypical Part Network (R3-ProtoPNet)}
\label{tab:r3-ppnet}

\begin{figure*}[ht]
  \includegraphics[width=\textwidth, trim = 4cm 20cm 3cm 16cm]{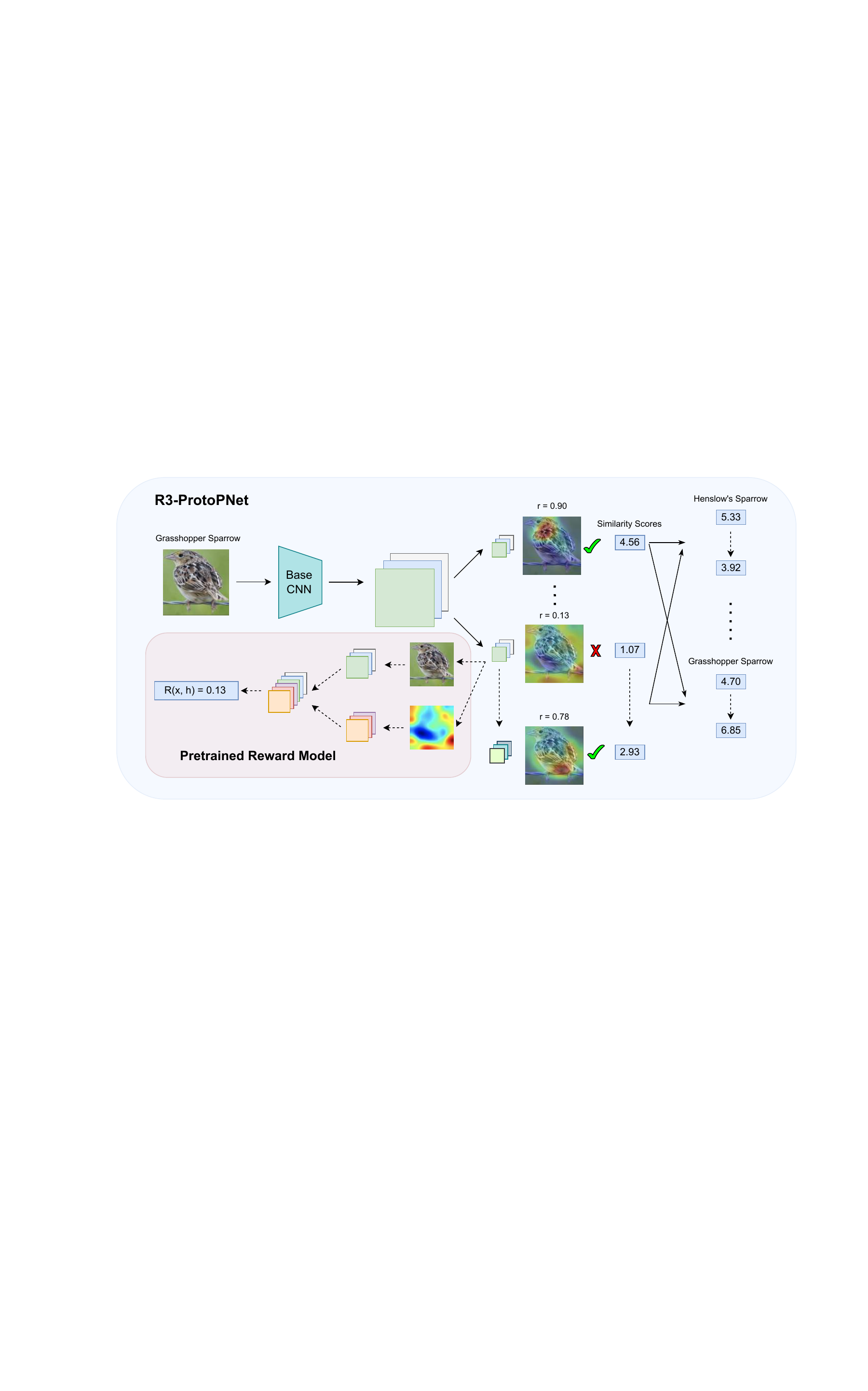}
\caption{An overview of the R3-ProtoPNet framework. Dashed arrows indicate our R3 debugging procedure.} \label{fig:framework}
\end{figure*}

After having collected high-quality human feedback data and trained a reward model, we can now incorporate it into a fine-tuning framework to improve the interpretability of ProtoPNet. We incorporate the reward model via a three step process consisting of reward weighing, reselection, and retraining. Each step is described in more detail below. Figure \ref{fig:framework} provides an overview of our proposed framework.

\subsubsection{Reward Reweighing} 

Although PPO is a popular option for RLHF \citep{ouyang2022training}, there is evidence that simpler fine-tuning algorithms can lead to similar performance increases \citep{askell2021general}. Inspired by the success and the efficiency of reward-weighted learning \citep{lee2023aligning, stiennon2022learning, ziegler2019fine}, we develop a straightforward reward-weighted update for the ProtoPNet:

\begin{equation}
\max_{p_j}\mathcal{L}_{reweigh}(z_i^*, p_j) = \max_{p_j} \sum_{i \in I(p_j)}^{n}\frac{r(x_i, p_j)}{\lambda_{dist}{\|z_i^* - p_j\|}^2_2 + 1}
\label{tab:eq1}
\end{equation}

where $z_i^*$ = $\text{argmin}_{z\in\text{patches}(f(x_i))}{\|z - p_j\|}_2^2$, \\$I(p_j)$ = $\{i\ \lvert\ y_i\in\text{class}(p_j)\}$, and $\lambda_{dist}$ is a fixed hyperparameter. We note that the objective function $\mathcal{L}_{reweigh}$ is a sum of the inverse distances weighted by the reward of the prototype on that image. Since we only update the prototype $p_j$, the only way to maximize this objective is to minimize the distance between prototype and image patches with high reward $r(x_i, p_j)$. This causes the prototype to resemble high reward image patches, improving the overall quality of the prototypes. Wanting to preserve prototypes that already have high reward, we only update those prototypes that have relatively low mean reward $\gamma$, and choose this reweighing threshold per base architecture and reward model (see the Appendix for threshold choices). $\lambda_{dist}$ is included in the objective function to rescale distances, since the closest distances are near zero. We find best performance with $\lambda_{dist} = 100$.

Practically, we find that optimizing this objective function leads to locally maximal solutions, resulting in local updates that do not modify prototypes with low quality values of $1$, but it's more likely to improve prototypes with quality values of $2$ or higher. If the prototype $p_j$ has high activation over the background of an image $x_i$, for example, the closest patches $z_i^*$ in the training data will also be background patches, and the reward of the prototype will be low, leaving minimal room for change. In other words, reward reweighing moves the prototype gradually toward better locations on the reward manifold, but it is not possible to dramatically change the location of the illuminated patch via this loss function. 

\subsubsection{Prototype Reselection}

In order to improve low quality prototypes that require significant manipulation, we introduce a reselection procedure based on a reward threshold. Given a prototype $p_j$ and image $x_i$, if $\frac{1}{n_k}\sum_{i\in I(p_j)}r(x_i, p_j) < \alpha$, where $\alpha$ is a pre-determined threshold and $n_k$ is the number of training images in class $k$, or if the patch of the given prototype $p_j$ matches the patch of another prototype of the same class, we reselect the prototype. The reselection process involves iterating over patch candidates $z'_i$ and temporarily setting the prototype $p'_j = z'_i$, where $z'_i$ is chosen randomly from the patches of a randomly selected image $x'_i$ in the class of $p_j$. If $\frac{1}{n_k}\sum_{i\in I(p_j)}r(x_i', p'_j) > \beta$, where $\beta$ is an acceptance threshold, and if none of the prototypes match patch $p'_j = z'_j$, then we accept the patch candidate as the new prototype. We found that varying the $\alpha$ and $\beta$ values per base architecture led to the best performance (See Appendix \ref{tab:thresh} for threshold choices). We refer to the combination of reweighing and reselection as the R2 update step, and the corresponding trained model the R2-ProtoPNet.

The reasoning process behind our prototype reselection method takes inspiration from the original push operation in \citet{chen2019looks}. Similar to how ProtoPNet projects prototypes onto a specific training image patch, here we reselect prototypes to be a particular reward-filtered training image patch. With a high enough acceptance threshold $\beta$, this forces the elimination of low reward prototypes while preserving the information gain of having an additional prototype.  

One possible alternative approach is to instead search over the training patches and select those patches with the highest reward, but we found that randomly selecting patches, in place of an exhaustive search, led to higher prototype diversity and less computation time. 

While we do not use a traditional reinforcement learning algorithm to fine-tune our model as is typically done in RLHF \citep{askell2021general}, pairing the reselection/reward-reweighing (R2 update) and retraining steps together resembles the typical explore-exploit trade-off in RL problems: the R2 update serves as a form of exploration, drastically increasing the quality of uninformative prototypes by breaking their dependence on spurious features, while retraining with the updated prototypes resembles exploit behavior, improving upon already high-quality prototypes. 

\subsubsection{Retraining}

A critical step missing in the R2 update is a connection to prediction accuracy. As discussed in Section \ref{experiments}, without incorporating predictive information, performing the reward update alone results in lowered test accuracy. Since the above updates only act on the prototypes themselves, not the rest of the network, the result is a misalignment between the prototypes and the model's base features and final classifier layer. The reward update guides the model towards more interpretable prototypes, but the reward update alone fails to use the higher quality prototypes for better prediction. 

To account for the lack of predictive performance, the final step of R3-ProtoPNet is retraining. With the updated prototypes, simply retraining using the same loss function used in the original ProtoPNet training results in the realignment of the prototypes and the rest of the model. Although one could worry that predictive accuracy would reduce the interpretability of the model \citep{rudin2021interpretable}, we find that retraining increases predictive accuracy while maintaining the quality increases of the R2 update. The result is a high accuracy model with higher-quality prototypes. We explore evidence of this phenomenon and why this is the case in the following section. 

\section{Experiments} \label{experiments}

\subsection{Bird Species Identification}
Here we discuss the results of training the R3-ProtoPNet on the CUB-200-2011 dataset, the same dataset as used in \citet{chen2019looks}. 
\subsubsection{Data Preprocessing}
R3-ProtoPNet requires the original dataset for the initial training, as well as additional scalar ratings of the selected activation patterns produced by image-prototype pairs. Combined, this results in the dataset described in Section \ref{rwr}. To offer better comparison against the original ProtoPNet, we use the same dataset for initial training that was used in \citet{chen2019looks}, the CUB-200-2011 dataset \citep{Wah2011TheCB}. The CUB-200-2011 dataset consists of roughly 30 images of 200 different bird species. We employ the same data augmentation scheme used in \citet{chen2019looks}, which adds additional training data by applying a collection of rotation, sheer, and skew perturbations to the images, resulting in a larger augmented dataset. 

For the collection of the activation pattern ratings, we only provided the activation patterns overlaid on the original images to the raters. Using Amazon Mechanical Turk to recruit six workers per prototype-image pair, we take the average as the user-provided rating for that pair. We also exclude the entries with $|r_{i,j} - r_{i',j'}| < 0.5$ to increase the contrast between pairs. In total, 700 rated prototype-image pairs are collected according to the scale approach described in Figure \ref{fig:scale}, and we randomly selected 500 of them (the rest were used as a held-out test set to evaluate the robustness) to train the reward model.

\subsubsection{Implementation}
\label{tab:implementation}

Similar to \citet{chen2019looks}, we study the performance of R3-ProtoPNet across five different base architectures: VGG-19, ResNet-34, ResNet-50, DenseNet-121, and DenseNet-161. While the original ProtoPNet sets the number of prototypes per class at $m_k = 10$, we additionally run the VGG-19 architecture with $m_k=5$ prototypes to explore model performance when the number of prototypes is limited. No other modifications were made to the original ProtoPNet architecture. At most 50 epochs are needed in this initial training step. 

The reward model $r(x_i, h_i)$ is similar to the base architecture of the ProtoPNet. Two ResNet-50 base CNNs take in the input image $x_i$ and the associated acticvation pattern $h_i$ separately, and both have two additional convolutional layers. The outputs of the convolutional layers are concatenated and fed into a final linear layer with sigmoid activation to predict the Bradley-Terry ranking. Predicted rewards are therefore bound in the range $(0, 1)$. We train the reward model for 5 epochs on a synthetic comparison dataset of 49K paired images and preference labels derived from 500 human ratings, and evaluate on 14K testing pairs. The reward model achieves 90.09\% test accuracy. We additionally analyze the sensitivity of the reward model and R3-ProtoPNet to the amount of human feedback used for reward model training (see Appendix \ref{sensitivity}), and the results suggest that the performance gain of R3-ProtoPNet can be achieved with even fewer human ratings (around 300 image-prototype pairs). 

\subsubsection{Evaluation Metrics}

To evaluate the performance of R3-ProtoPNet, we compare it to ProtoPNet on three metrics: test accuracy, reward, and activation precision (AP). We use test accuracy to measure the predictive performance of the models. As the above section demonstrates, the learned reward model achieves high accuracy in predicting which prototype ranks above another in accordance with human preferences, so we therefore use it as a measure of prototype quality. The final metric, activation precision, is a common metric that has been used in prior work to evaluate the overlap between a prototype's activations and the pixels associated with a given bird \citep{barnett2021iaiabl}, which provides another metric of interpretability independent of our method. In our work, we report a modified version of AP introduced in \cite{bontempelli2023conceptlevel} to consider the specific value of the activation at each single pixel, not just the overlap alone.

\subsubsection{Results}

\begin{table*}[hbt!]
\caption{R3 updates tend to increase the test accuracy. Average accuracies and standard deviations are reported across five runs, where $m_k$ is the number of prototypes per class. }
\label{tab:accuracies}
\vskip 0.15in
\begin{center}
\begin{small}
\begin{sc}
\begin{tabular}{lccc}
\toprule
\textbf{Base ($m_k)$} & \textbf{ProtoPNet} & \textbf{R2-ProtoPNet} & \textbf{R3-ProtoPNet}\\
\midrule
VGG-19 ($5$)        & $76.33 \pm 0.12$ & $62.76 \pm 1.18$ & $\mathbf{77.80} \pm 0.18$  \\
VGG-19 ($10$)       & $77.58 \pm 0.22$ & $50.41 \pm 1.36$ & $\mathbf{79.60} \pm 0.25$ \\  
ResNet-34 ($10$)    & $78.73 \pm 0.13$ & $58.11\pm 2.71$ & $\mathbf{80.21} \pm 0.22$ \\
ResNet-50 ($10$)    & $78.52 \pm 0.17$ & $56.36\pm 2.40$  & $\mathbf{80.25} \pm 0.22$  \\
DenseNet-121 ($10$) & $79.64 \pm 0.23$ & $54.67 \pm 2.29$   & $\mathbf{80.42} \pm 0.26$  \\
DenseNet161  ($10$) & $\mathbf{79.75} \pm 0.27$ & $62.75 \pm 2.43$ & $79.48 \pm 0.36$  \\
Ensemble of Above   & $82.92 \pm 0.09$ & $70.46 \pm 0.82$  & $\mathbf{84.37} \pm 0.20$ \\
\bottomrule
\end{tabular}
\end{sc}
\end{small}
\end{center}
\end{table*}

\begin{table*}[hbt!]
\caption{R3-ProtoPNet outperforms the original ProtoPNet in terms of the image-prototype rewards estimated by our reward model. Values are averaged over the entire test dataset. We divide the R2 update into two columns \textbf{Reselected} and \textbf{Reweighed} to better show individual effect of each step. We omit the standard deviations as the values are small.}
\label{tab:rewards}
\vskip 0.15in
\begin{center}
\begin{small}
\begin{sc}
\begin{tabular}{lcccc}
\toprule
\textbf{Base ($m_k)$} & \textbf{ProtoPNet} & \textbf{Reselected} & \textbf{Reweighed} & \textbf{R3-ProtoPNet}\\
\midrule
VGG19 ($5$) &  0.61 & 0.66 & 0.70 & 0.71 \\

VGG19 ($10$) & 0.46 & 0.55 & 0.64 & 0.67 \\

ResNet-34 ($10$)  & 0.40 & 0.47 & 0.51 & 0.54 \\

ResNet-50 ($10$)  & 0.36 & 0.45 & 0.50 & 0.54 \\

DenseNet-121 ($10$) & 0.48 & 0.53 & 0.58 & 0.58 \\

DenseNet-161 ($10$) & 0.48 & 0.51 & 0.57 & 0.56 \\

Average & 0.47 & 0.53 & 0.58 & 0.60 \\
\bottomrule

\end{tabular}
    
\end{sc}
\end{small}
\end{center}
\end{table*}

\begin{table*}[hbt!]
\caption{Average Activation Precision (AP) over the test dataset are increased across different stages of R3 updates. }
\label{tab:ap}
\vskip 0.15in
\begin{center}
\begin{small}
\begin{sc}
\begin{tabular}{lcccccr}
\toprule
\textbf{Base ($m_k)$} & \textbf{ProtoPNet} & \textbf{Reselected} & \textbf{Reweighed} & \textbf{R3-ProtoPNet}\\
\midrule
VGG19 ($5$) &  $70.31 $ & $79.81 $ & $85.64 $ & $86.61 $ \\
VGG19 ($10$) & $63.12 $  & $75.95 $ & $82.72 $ & $81.62 $\\
ResNet-34 ($10$)  & $85.63 $ & $88.81 $ & $90.33 $  & $92.23 $\\
ResNet-50 ($10$)  & $71.45 $ & $79.29 $ & $83.69 $  & $83.52 $\\
DenseNet-121 ($10$) & $66.22 $ & $81.64 $ & $86.73 $  & $89.38 $ \\
DenseNet-161 ($10$) & $82.56 $ & $85.24 $ & $87.55 $  & $87.60 $ \\
Average & 73.22 & 81.79 & 86.11 & 86.83\\
\bottomrule

\end{tabular}
\end{sc}
\end{small}
\end{center}
\end{table*}

After training ProtoPNet, running the R2 update step, and then performing retraining, we see several trends across multiple base architectures. In Table \ref{tab:accuracies}, we report the test accuracy of the different base architectures across stages of R3-ProtoPNet training. Generally, the test accuracy from ProtoPNet temporarily decreases after applying the R2 update, but retraining could effectively recover the predictive loss, in most cases notably improving test accuracy.  

In Table \ref{tab:rewards} and Table \ref{tab:ap}, we report the average reward and the activation precision metrics. Compared with ProtoPNet, R3-ProtoPNet increases the average reward and activation precision across all prototypes, test images, and base architectures by $27.66\%$ and $18.59\%$, respectively. Here we note that the average reward and AP serve as complementary interpretability metrics, as in a single stage the reward and AP values across different base architectures could have different patterns - this is because AP only considers the overlap between top 5\% activated regions and the bird body, while the reward model/human user takes into account a much larger activated regions with warm colors. However, the increasing patterns for each architecture across different R3 stages are highly consistent, which demonstrate R3's success in aligning with human preference and improving model interpretability. 


\subsubsection{Discussion}

\begin{figure}[t]
  \includegraphics[width=0.5\textwidth, trim = 5.5cm 18cm 5.5cm 5cm]{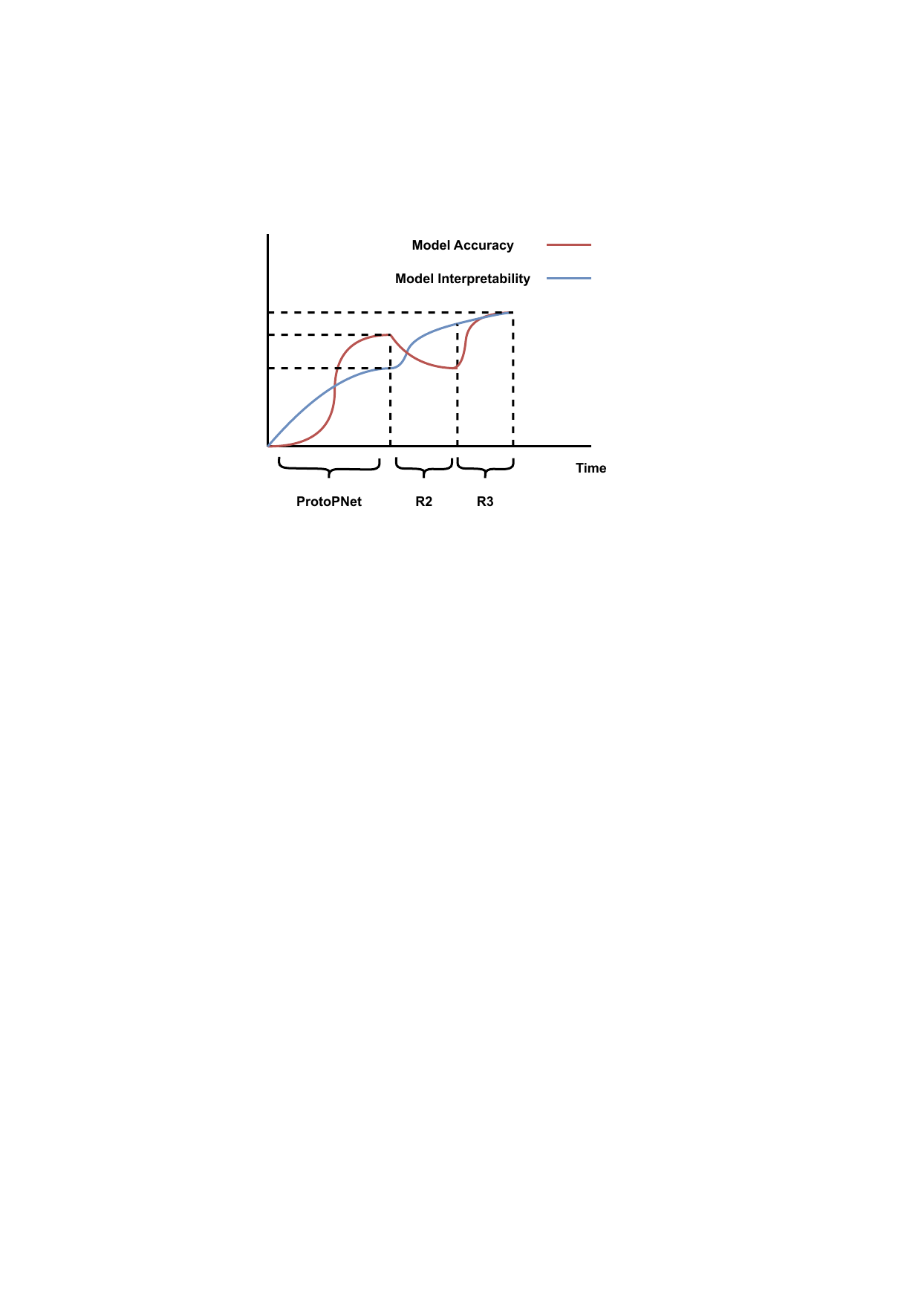}
\caption{Trade-off curves between model accuracy and model interpretability. The plot is qualitative.} \label{fig:tradeoff}
\end{figure} 

Given the above results, we can observe that although test accuracy experiences a substantial drop during the R2 update, when both reward and AP increase significantly, the model's predictive power is later restored after retraining and in most cases even further improved compared to the original ProtoPNet. On the other hand, the retraining stage doesn't hurt either reward or AP, but instead result in a slight increase of both. This phenomenon suggests that there doesn't exist a long-term trade-off between accuracy and interpretability: this trade-off only temporarily occurs when the spurious feature attributions haven't been removed by our debugging procedure; and once those predictive short-cuts (which tends to have some but limited predictive power) are detected and eliminated, there should be a positive correlation between predictive accuracy and model interpretability. Figure \ref{fig:tradeoff} illustrates this empirical trade-off across different ProtoPNet pretraining and R3 debugging stages. 

\subsection{Car Model Identification}
In addition to bird species classification, we also conduct experiments on the Stanford Cars dataset \citep{stanford-cars}. This dataset contains 196 different classes with a train/test split of 8144/8041 images. With the implementation remains the same, we found that R3-ProtoPNet outperforms ProtoPNet in a very similar way to the CUB-200-2011 dataset. We include the empirical results of test accuracies and average rewards in Appendix \ref{tab:stanford-cars}.

\section{Generalizability of the R3 Framework} 

To test whether the debugging effects of our R3 framework on ProtoPNet could generalize to other models, we incorporate R3 into two other recent models ProtoPFormer \citep{xue2022protopformer} and ProtoPNet with shallow-deep feature alignment (SDFA) and score aggregation (SA) modules \citep{evalprotopnet}. With similar experiment setup, we find that although these two models already perform better than ProtoPNet before debugging, our R3 framework is still able to slightly improve them both in terms of accuracy and interpretability. This suggests that our R3 framework could bring forth incremental performance gain to other prototype-based variants. The detailed experiment results can be found in Appendix \ref{tab:variants}.

\section{Limitations and Future Work}\label{discussion}

While R3 succeeds in bringing forth interpretability and predictive performance gains, there's still room for improvement. For example, the reward model is trained on ratings of individual image-prototype pairs, mostly focusing on overlap and single-image consistency (i.e. whether the prototype simultaneously focuses on multiple body parts in that image), while ignoring cross-image preferences, such as whether the prototype focuses on different parts across images. We also note that R3-ProtoPNet fails to completely eliminate duplicate prototypes, with several high-reward prototypes converge to the same part of the image. To address these issues, it's promising to extend ratings to multiple image-prototype pairs and create more diverse reward models, possibly using them in ensemble. Meanwhile, similar to many other human preference learning scenarios, our ratings to be collected also rely on certain level of subjective individual judgement calls, and this would inevitably lead to noises in the reward labels, which should be further minimized, according to the \textit{predictability, computability, and stability} (PCS) framework for veridical data science \citep{yu2020veridical}.

Another limitation with R3-ProtoPNet and other methods that rely on human feedback is that the model itself might be learning features that, while seemingly confusing to a human, are helpful and meaningful for prediction. \citet{barnett2021iaiabl} argue that the ProtoPNet can predict with non-obvious features like texture and contrast, which might be penalized via a learned reward function. An interesting line of future work is to investigate how certain ProtoPNet variants could critique human feedback, and argue against a human-biased reward model.

While this work focuses on improving the performance of ProtoPNet, a major benefit of reward-based finetuning is its flexibility in application. With proper adaptations, we expect our R3 debugging framework could generalize to many other prototype-based or interpretable machine learning models and serve as a useful concept-level debugging tool.

\section{Conclusion}

In this work, we present the R3 debugging framework, an efficient and generalizable human-in-the-loop approach to improve the class of prototype-based deep learning models. Our work is the first method that uses a learned reward model to quantify the qualitative prototypical visual explanations and use them to improve the model itself. Our experiments show both increased model performance and improved model interpretability. It's demonstrated by our work that the ability of reward learning to quantify qualitative human preferences make reward-based fine-tuning a promising direction for the improvement of interpretable deep models.

\section*{Acknowledgements} 
We would like to acknowledge the partial support from an MSR grant through BAIR at UC Berkeley, NSF grant 2023505 on Collaborative Research: Foundations of Data Science Institute (FODSI), and NSF grant MC2378 to the Institute for Artificial CyberThreat Intelligence and OperatioN (ACTION).

\section*{Impact Statement}
Our proposed R3 framework is an effective debugging tool for interpretable prototype-based neural networks in the field of Computer Vision, and users could incorporate it into other existing models to improve their performance, thus making more accurate and interpretable decisions. On the other hand, users can use pretrained reward models to evaluate the explanation quality of the models, so they will have a more informed decision on whether to trust the model output.  

Furthermore, our framework enables users to train reward models according to their own preferences, which could make the models interpretable in a personalized way. 

With all the reasons above, we believe our work will have positive impact on the machine learning and interpretable deep learning communities.

\bibliography{main}
\bibliographystyle{icml2024}

\clearpage

\newpage
\appendix
\onecolumn

\section{Pairwise Loss Function for the Reward Model}
\label{tab:pairwise-loss}
For completeness, here is the explicit formulation of the loss function described in Section 4.2:
\begin{equation*}
\begin{aligned}
\mathcal{L}_{\text{reward}} = & -\sum_{i\neq i' \text{ or } j\neq j'} \bigg[\mathbf{1}_{c_{iji'j'} = -1} \log\left(\frac{\exp(r(x_i, h_{ij}))}{\exp(r(x_i, h_{ij})) + \exp(r(x_{i'}, h_{i'j'}))}\right) \\
& + \mathbf{1}_{c_{iji'j'} = 1} \log\left(\frac{\exp(r(x_{i'}, h_{i'j'}))}{\exp(r(x_i, h_{ij})) + \exp(r(x_{i'}, h_{i'j'}))}\right)\bigg]
\end{aligned}
\end{equation*}
where $c_{i, j, i', j'}$ refers to the comparison value associated with the column indexed by $i,j,i',j'$ in the synthetic dataset $\mathcal{D}_{paired}$, which is explained in section \ref{reward-learning}. The architecture of the reward model is detailed in section \ref{tab:implementation}.

\section{Thresholds and the Number of Updated Prototypes}
\label{tab:thresh}
As described in Section \ref{tab:r3-ppnet}, for each base architecture, the various thresholds for reweighing, reselection, and acceptance. These thresholds were chosen by examining the reward distribution of the base architectures to see if prototypes with low reward cluster around any particular values. Across models, a reweighing threshold of $0.4$ or $0.35$ sufficed, but further tuning was needed for the reselection and acceptance thresholds. We present the final thresholds used for each R2 step in Table \ref{tab:th-values}. 

\begin{table*}[hbt!]
\caption{Thresholds used across base architectures during R2 step for the CUB dataset.}
\label{tab:th-values}
\vskip 0.15in
\begin{center}
\begin{small}
\begin{sc}
\begin{tabular}{lcccc}
\toprule
\textbf{Base ($m_k)$} & Reselection Threshold & Reweigh Threshold & Acceptance Threshold \\
\midrule
VGG-19 (5) & 0.35 & 0.40 & 0.50 \\
VGG-19 (10) & 0.25 & 0.40 & 0.43 \\
ResNet-34 (10) & 0.22 & 0.35  & 0.40 \\
ResNet-50 (10) & 0.18 & 0.35 & 0.40 \\
DenseNet-121 (10) & 0.25 & 0.35 & 0.45  \\
DenseNet-161 (10) & 0.25 & 0.35  & 0.43 \\
\bottomrule

\end{tabular}
\end{sc}
\end{small}
\end{center}
\end{table*}

Using the reselection thresholds above, we report the total number of updated prototypes for each architecture in Table \ref{tab:updated}.  

\begin{table*}[hbt!]
\caption{Total number of prototypes updated across the two R2 steps, divided by the total number of prototypes for that network.}
\vskip 0.15in
\begin{center}
\begin{small}
\begin{sc}
\label{tab:updated}
\begin{tabular}{lcc}
\toprule
\textbf{Base ($m_k)$} & \textbf{\#Reselected Prototypes} & \textbf{\#Reward-Reweighed Prototypes} \\
\midrule
VGG-19 (5) & 107 / 1000 & 432 / 1000 \\
VGG-19 (10) & 384 / 2000 & 644 / 2000 \\
ResNet-34 (10) & 349 / 2000 & 702 / 2000 \\
ResNet-50 (10) & 365 / 2000 & 749 / 2000 \\
DenseNet-121 (10) & 294 / 2000 & 662 / 2000 \\
DenseNet-161 (10) & 276 / 2000 & 598 / 2000 \\
\bottomrule

\end{tabular}
\end{sc}
\end{small}
\end{center}
\end{table*}

\section{Sensitivity to Amount of Human Feedback}
\label{sensitivity}
To evaluate the influence of the amount of human feedback on our R3 framework, we experiment using fewer human ratings to train a reward model and then perform the R3 updates. The results are in Table \ref{tab:sensitivity}. Although we used 500 ratings to reach the peak performance in the main experiments, it's observed that the R3 framework starts to improve the original ProtoPNet when the number of collected ratings reaches 300.

\begin{table}[hbt!]
\caption{Test accuracies of the trained reward models and the debugged R3-ProtoPNets (ensembled) given different amount of human ratings (CUB dataset).}
    \centering
    \begin{tabular}{|c|c|c|c|c|c|}
        \hline
        \diagbox{Metric}{\#Ratings} & \textbf{100 ratings} & \textbf{200 ratings} & \textbf{300 ratings} & \textbf{400 ratings} & \textbf{500 ratings}\\
        \hline
        Reward Model Acc. & $69.50 \pm 0.25$ & $77.34 \pm 0.25$ & $83.27 \pm 0.27$ & $88.67 \pm 0.16$ & $90.09 \pm 0.20$ \\
        \hline
        R3-ProtoPNet Acc. & $77.69 \pm 0.30$ & $80.15 \pm 0.22$ & $83.06 \pm 0.28$ & $84.03 \pm 0.21$ & $84.37 \pm 0.20$ \\
        \hline

    \end{tabular}
    
    \label{tab:sensitivity}
\end{table}

\section{Performance of R3-ProtoPNet on Stanford Cars Dataset}
\label{tab:stanford-cars}

Here we report the test accuracies and average rewards of R3-ProtoPNet on the Stanford Cars dataset in Table \ref{tab:cars-accuracy} and Table \ref{tab:cars-reward}. Note that although the reward values for this dataset tend to be higher than that of CUB-200-2011, which is due to the fact that we need to train a new reward model for this new dataset, the general trends are the same. We don't include the activation precision result because a fine-grained segmentation mask for this dataset is not available. 

\begin{table*}[hbt!]
\caption{R3 updates tend to increase the test accuracy for the Stanford Cars dataset. Average accuracies and standard deviations are reported across five runs, where $m_k$ is the number of prototypes per class. }
\label{tab:cars-accuracy}
\vskip 0.15in
\begin{center}
\begin{small}
\begin{sc}
\begin{tabular}{lccc}
\toprule
\textbf{Base ($m_k)$} & \textbf{ProtoPNet} & \textbf{R2-ProtoPNet} & \textbf{R3-ProtoPNet}\\
\midrule
VGG-19 ($5$)        & $85.10 \pm 0.15$ & $69.61 \pm 1.23$ & $\mathbf{86.75} \pm 0.18$  \\
VGG-19 ($10$)       & $87.25 \pm 0.18$ & $66.11 \pm 2.10$ & $\mathbf{88.71} \pm 0.27$ \\  
ResNet-34 ($10$)    & $85.62 \pm 0.08$ & $58.73\pm 2.36$ & $\mathbf{87.18} \pm 0.14$ \\
ResNet-50 ($10$)    & $85.27 \pm 0.21$ & $62.66\pm 2.77$  & $\mathbf{87.25} \pm 0.19$  \\
DenseNet-121 ($10$) & $86.03 \pm 0.18$ & $63.49 \pm 1.88$ & $\mathbf{86.59} \pm 0.25$  \\
DenseNet161  ($10$) & $88.19 \pm 0.29$ & $60.75 \pm 2.23$ & $\mathbf{89.48} \pm 0.31$  \\
Ensemble of Above   & $90.39 \pm 0.14$ & $73.42 \pm 0.81$  & $\mathbf{91.57} \pm 0.24$ \\
\bottomrule
\end{tabular}
\end{sc}
\end{small}
\end{center}
\end{table*}

\begin{table*}[hbt!]
\caption{Average rewards during different R3 debugging stages (Stanford Cars dataset).}
\label{tab:cars-reward}
\vskip 0.15in
\begin{center}
\begin{small}
\begin{sc}
\begin{tabular}{lcccc}
\toprule
\textbf{Base ($m_k)$} & \textbf{ProtoPNet} & \textbf{Reselected} & \textbf{Reweighed} & \textbf{R3-ProtoPNet}\\
\midrule
VGG19 ($5$) & 0.78  & 0.87 & 0.91 & 0.92 \\
VGG19 ($10$) & 0.73 & 0.82 & 0.87 & 0.87 \\
ResNet-34 ($10$) & 0.69 & 0.75 & 0.82 & 0.85 \\
ResNet-50 ($10$) & 0.66 & 0.74 & 0.79 & 0.81 \\
DenseNet-121 ($10$) & 0.75 & 0.80 & 0.83 & 0.86 \\
DenseNet-161 ($10$) & 0.72 & 0.80 & 0.82 & 0.84 \\
Average & 0.72 & 0.80 & 0.84 & 0.86 \\
\bottomrule

\end{tabular}
    
\end{sc}
\end{small}
\end{center}
\end{table*}

\section{Debugging Effects of R3 on Other ProtoPNet Variants}
\label{tab:variants}
To test the generalizability of our R3 framework, we first apply our R3 framework to ProtoPFormer \citep{xue2022protopformer}, which is another ProtoPNet extension that uses vision transformer (ViT) backbones. In ProtoPFormer, two types of prototypes are used: the global prototypes are able to provide holistic views of the objects and eliminate confounding effects of the background, while the local prototypes capture the fine-grained visual features that are useful for classification. Empirically we found success applying our R3 framework to update both global and local prototypes. The results are summarized in Table \ref{tab:protopformer}.

\citet{evalprotopnet} propose to improve the stability and consistency of the original ProtoPNet by adding 1) a shallow-deep feature alignment (SDFA) module, which helps preserve the spatial information of deep feature maps by incorporating spatial information from shallow layers into deep layers, and 2) a score aggregation (SA) module, which improves the model by aggregating activation values only into corresponding categories. We apply our R3 framework to this augmented ProtoPNet, and the results are included in Table \ref{tab:evalprotopnet}.  

These two experiments show that the R3 debugging procedure could generalize well toward other prototype-based models.

\begin{table*}[hbt!]
\caption{Performance report of R3-ProtoPFormer across different stages. We used the best-performing DeiT-S backbone.}
\label{tab:protopformer}
\vskip 0.15in
\begin{center}
\begin{small}
\begin{sc}
\begin{tabular}{lcccc}
\toprule
\textbf{Metric} & \textbf{ProtoPFormer} & \textbf{Reselected} & \textbf{Reweighed} & \textbf{R3-ProtoPFormer}\\
\midrule
Test Accuracy & $84.27 \pm 0.20$ & $73.35 \pm 1.52$ & $71.79 \pm 2.13$ & $\mathbf{85.31} \pm 0.23$ \\
Average Reward (Global) & 0.55 & 0.62 & 0.66 & 0.68 \\
Average Reward (Local) & 0.59 & 0.62 & 0.67 & 0.70 \\
Activation Precision (Global) & 83.48 & 86.59 & 87.96 & 87.43 \\
Activation Precision (Local)  & 86.63 & 88.28 & 89.11 & 89.36 \\

\bottomrule

\end{tabular} 
\end{sc}
\end{small}
\end{center}
\end{table*}

\begin{table*}[hbt!]
\begin{center}
\caption{Performance report of debugging the augmented ProtoPNet (with SDFA and SA modules), across different R3 stages. We used the best-performing DenseNet-161 backbone.}
\label{tab:evalprotopnet}
\vskip 0.15in
\begin{small}
\begin{sc}
\begin{tabular}{lcccc}
\toprule
\textbf{Metric} & \textbf{ProtoPNet with SDFA and SA} & \textbf{Reselected} & \textbf{Reweighed} & \textbf{Retrained}\\
\midrule
Test Accuracy & $85.03 \pm 0.14$ & $75.30 \pm 1.82$ & $72.89 \pm 2.05$ & $\mathbf{85.82} \pm 0.22$ \\
Average Reward & 0.68 & 0.70 & 0.74 & 0.74 \\
Activation Precision & 89.21 & 90.50 & 92.39 & 92.88 \\

\bottomrule

\end{tabular} 
\end{sc}
\end{small}
\end{center}
\end{table*}

\section{Prototype Examples}

\begin{figure}[hbt!]
  \includegraphics[width=\textwidth]{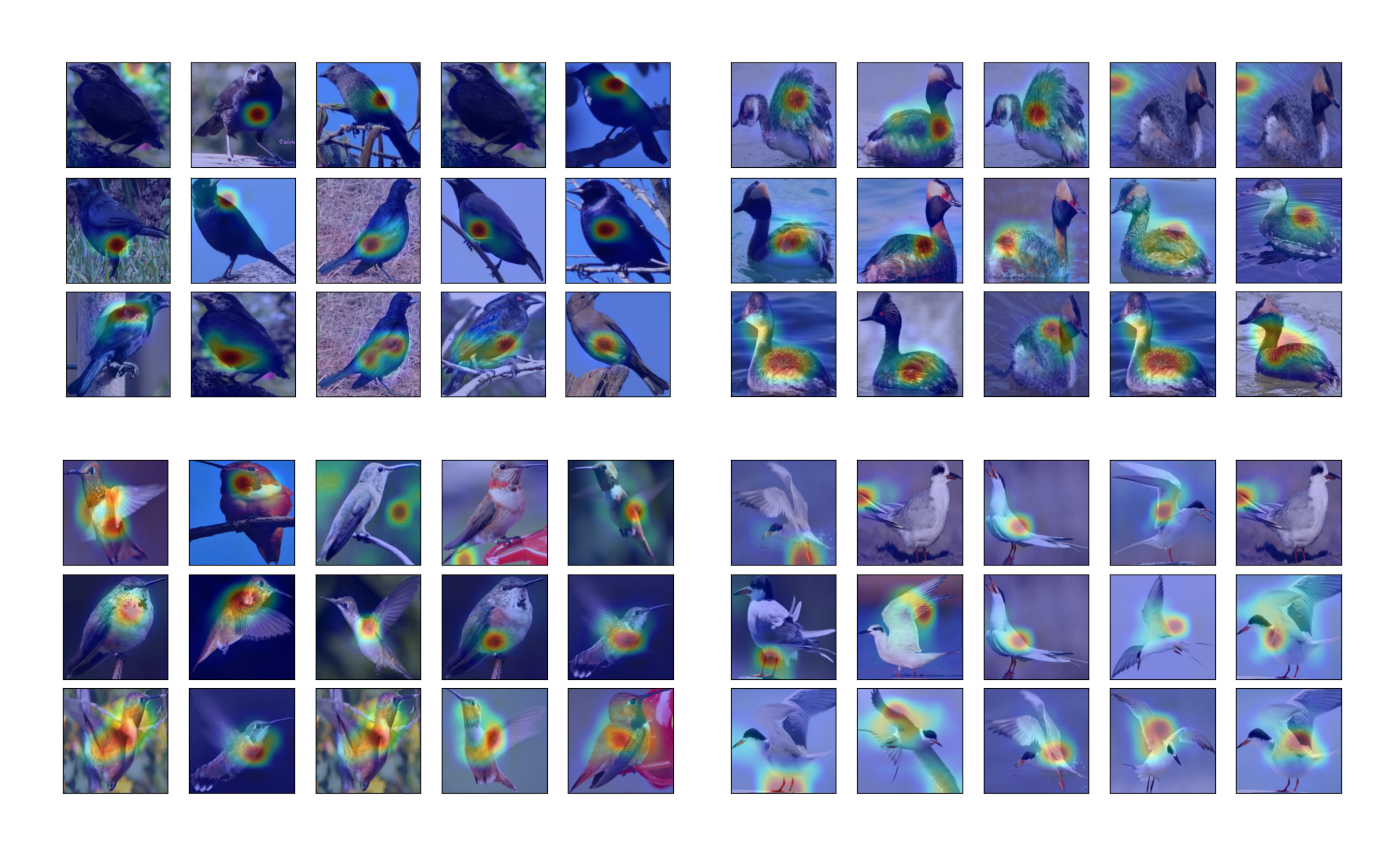}
\caption{Closest training patches of the five prototypes of ProtoPNet (top row), R2-ProtoPNet (middle row), and R3-ProtoPNet (bottom row) within the same class (5 prototypes per class). Each cluster of 3 rows of images is a seperate class.} \label{fig:proto1}
\end{figure}

Here we provide some examples of prototypes from ProtoPNet, R2-ProtoPNet, and R3-ProtoPNet. In Figure \ref{fig:proto1}, we visualize the changes of the first 5 prototypes within 4 different classes across different stages, by overlaying each prototype on its closest training image patch. A typical trend we observe is that the R2 update indeed centers otherwise off prototypes on the bird in question, and then the complete R3 update makes those prototypes helpful for prediction.

\begin{figure}[hbt!]
  \includegraphics[width=\textwidth]{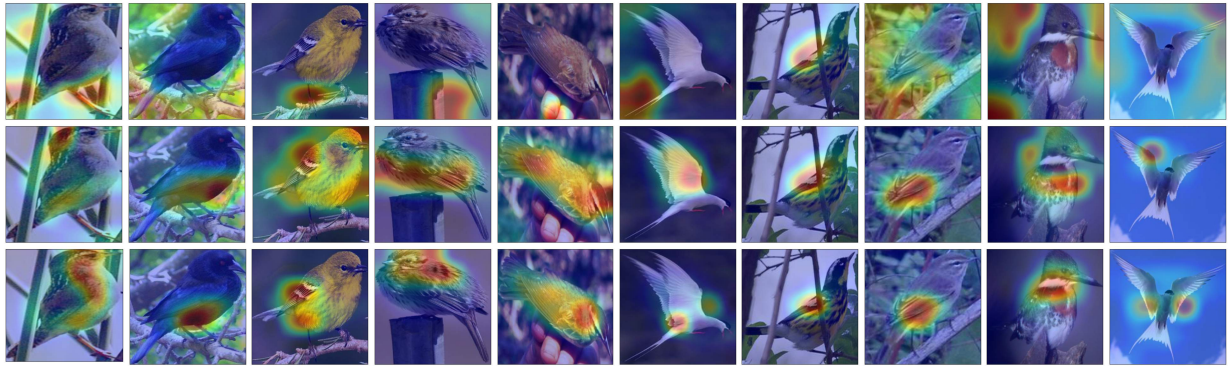}
\caption{Prototype projections on the same image (each column) from ProtoPNet (top row), R2-ProtoPNet (middle row), and R3-ProtoPNet (bottom row).} \label{fig:proto2}
\end{figure}

In Figure \ref{fig:proto2}, we focus on individual image-prototype pairs. Each column illustrates how the prototype changes during the R2 and R3 update when the image is held fixed. We see that the initially low-quality prototypes are gradually corrected, finally without having much dependence on spurious features like the background.
  
\end{document}